%% file: main.tex
\documentclass[letterpaper, 10 pt, conference]{ieeeconf}  % Comment this line out if you need a4paper

\IEEEoverridecommandlockouts                              % This command is only needed if 
\usepackage{amsmath,amsfonts}
\usepackage{array}
\usepackage{textcomp}
\usepackage{stfloats}
\usepackage{url}
\usepackage{verbatim}
\usepackage{cite}
\usepackage{times}

\usepackage{multicol}
\usepackage[bookmarks=true]{hyperref}
\hypersetup{colorlinks = true}
\usepackage{graphics} % for pdf, bitmapped graphics files
\usepackage{epsfig} % for postscript graphics files
\usepackage{amssymb}  % assumes amsmath package installed
\usepackage{bm}
\usepackage{subcaption}
\usepackage{array}
\usepackage{booktabs}
\usepackage{multirow}
\usepackage{makecell}
\usepackage{pifont}%
\usepackage[font={small,it}]{caption}

\usepackage{xcolor}
\usepackage{soul}

\title{\LARGE \bf
Real-Time Indoor Object SLAM with LLM-Enhanced Priors
}

\author{ Yang Jiao\textsuperscript{1}, Yiding Qiu\textsuperscript{1} and Henrik I. Christensen\textsuperscript{2}
\thanks{\textsuperscript{1}Students of Contextual Robotics Institute, University of California San Diego, La Jolla, CA 92093, USA \{\texttt{y4jiao, yiqiu}\}\texttt{@ucsd.edu}}% <-this % stops a space
\thanks{\textsuperscript{2}Henrik I. Christensen is with Faculty of the Department of Computer Science and Engineering, University of California San Diego, La Jolla, CA 92093, USA \texttt{hichristensen@ucsd.edu}}
}

\begin{document}

\maketitle
\thispagestyle{empty}
\pagestyle{empty}

%%%%%%%%%%%%%%%%%%%%%%%%%%%%%%%%%%%%%%%%%%%%%%%%%%%%%%%%%%%%%%%%%%%%%%%%%%%%%%%%

\begin{abstract}

Object-level Simultaneous Localization and Mapping (SLAM), which incorporates semantic information for high-level scene understanding, faces challenges of under-constrained optimization due to sparse observations. Prior work has introduced additional constraints using commonsense knowledge, but obtaining such priors has traditionally been labor-intensive and lacks generalizability across diverse object categories. We address this limitation by leveraging large language models (LLMs) to provide commonsense knowledge of object geometric attributes, specifically size and orientation, as prior factors in a graph-based SLAM framework. 
These priors are particularly beneficial during the initial phase when object observations are limited. 
We implement a complete pipeline integrating these priors, achieving robust data association on sparse object-level features and enabling real-time object SLAM. Our system, evaluated on the TUM RGB-D and 3RScan datasets, improves mapping accuracy by 36.8\% over the latest baseline. Additionally, we present real-world experiments in the supplementary video, demonstrating its real-time performance.

\end{abstract}

% We present a novel approach to Object SLAM that addresses the challenges of sparse observations in dynamic environments. Our system leverages Large Language Models (LLMs) to provide commonsense knowledge about object attributes, specifically size and orientation, as priors for mapping. These priors are particularly valuable during the initial observation phase when limited frames are available. By modeling these priors as factors in a graph-based SLAM framework, we effectively constrain the solution space for object pose estimation. Unlike traditional methods requiring extensive human annotation, our LLM-enhanced approach automatically generates priors for diverse object categories. We implemented a complete pipeline that combines these priors with a robust data association method, enabling real-time object SLAM. Evaluations on TUM RGB-D and 3RScan datasets demonstrate that our system significantly improves mapping accuracy and robustness in scenarios with limited inter-frame overlap.

%%%%%%%%%%%%%%%%%%%%%%%%%%%%%%%%%%%%%%%%%%%%%%%%%%%%%%%%%%%%%%%%%%%%%%%%%%%%%%%%

\section{Introduction}
\label{sect:intro}

Object Simultaneous Localization and Mapping (SLAM) builds environment maps by identifying and localizing objects, and using this information to infer the robot's position. Unlike traditional feature-based SLAM, object-level representations are sparse, focusing on semantic object data. Comparing to semantic segmentation on dense representations, such sparsity improves computational efficiency and reduces storage requirements. Meanwhile, high-level semantic information makes object SLAM useful for downstream tasks, such as 3D semantic scene reconstruction, object-goal navigation, and object retrieval \cite{wu2023object}.

When navigating an unfamiliar environment to find specific objects, the agent must efficiently build a map while moving quickly to cover large areas, often resulting in limited observations of objects from different viewpoints \cite{zhou2023efficient}. SLAM with sparse observations commonly occurred in real-world applications, such as when the camera frame rate is low relative to the agent's motion speed or when loop closure is unavailable. However, due to the inherent sparsity of object-level features, fewer-constraint object SLAM can be extremely challenging. Such sparsity affects the quality of initial mapping and might hinder subsequent optimization and data association. 
%It is also crucial in dynamic environments, such as grocery stores and homes, where the configuration of small objects changes frequently over time, necessitating frequent re-mapping \cite{doe-slam}.

% The performance of Kalman filter-based methods can be severely compromised due to the inter-frame discontinuity.
% TODO filter-based SLAM cannot work due to the discontinuity 
% Graph-based SLAM also faces the challenge of incremental optimization.
%
% Such sparsity may not be an impediment when using batch optimization \cite{nicholson2018quadricslam}, but the SLAM performance with incremental optimization can be compromised, especially with noisy observations at the early stage.

% It often collects only a few overlapping observations. This situation can occur when the camera frame rate is low or when loop closure is unavailable. In dynamic environments—like grocery stores or homes—object configurations change frequently, which calls for constant re-mapping.

%By leveraging typical object attributes, such as size and orientation, we narrow the solution space, thereby improving mapping accuracy and robustness even when observations are limited.
To address these issues, we introduce additional constraints from commonsense knowledge. Objects typically exhibit common attributes, such as size, shape, and orientation, which provide valuable priors for mapping systems. These attributes are essential during the initial observation phase, where limited 2D features and uncertain pose estimations bring challenges. By incorporating these priors, the solution space for object estimation can be effectively constrained, enhancing mapping accuracy in sparse observation scenarios. However, acquiring comprehensive object priors has traditionally been labor-intensive, requiring substantial human annotation across diverse object categories~\cite{jablonsky2018orientation}. Furthermore, commonsense knowledge about objects can vary by context, leading to potential mismatches between expected and actual observations. %These issues have impeded the broader application of prior-assisted object SLAM systems.

\begin{figure}[tbp]
    \setlength{\belowcaptionskip}{-10pt} 
    % \centering
    \centerline{\includegraphics[width=0.43\textwidth]{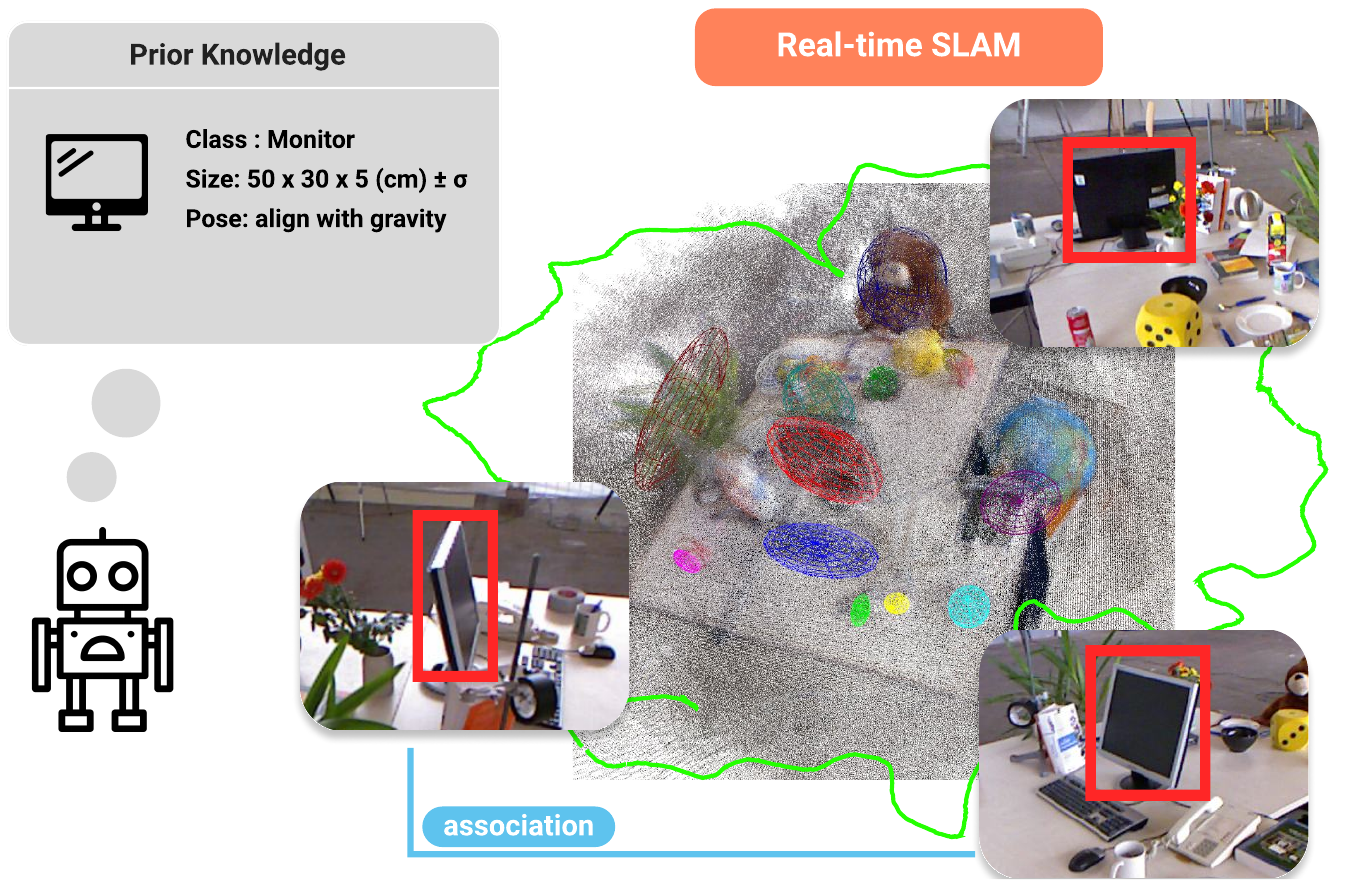}}
    \caption{%
    An overview of our object SLAM. We utilized prior knowledge of the objects from LLM and achieved a robust real-time object SLAM system. This approach enables the construction of a semantically meaningful, object-level map.
    }
    \label{fig:title-fig}
\end{figure}

% TODO citing fewer-frame paper, read their intro and associated references
% (MOVE TO LITERATURE REVIEW) Existing literature explores XYZ (certain aspects for fewer-frame SLAM).

We leveraged large language models (LLMs) to provide prior information, as commonsense inference is LLM's strong suit \cite{ge2024commonsense}.
We propose a LLM-enhanced prior-assisted object SLAM framework to address the challenge of under-constrained optimization in object SLAM.
Our approach models the prior knowledge from LLM as factors using a factor graph \cite{factorgraph910572}. Specifically, we focus on the geometric intrinsics of size and orientation for various object categories.
% Our approach uses object priors from two distinct types of information sources, and the priors are modeled as factors using a factor graph \cite{factorgraph910572}. The first type leverages commonsense knowledge, using size and orientation as key factors. The second type draws on observation priors, focusing on the object's centroid location using camera depth information. 
% The observation prior is designed because the initial object tracking often depends on an object's first few 2D detections, leading to potential inaccuracies in 3D projection.  
% TODO introduce LLM prior

% TODO rewrite summary
Our method integrates prior information from LLM and a combined data association method to achieve a full online object SLAM (Fig. \ref{fig:title-fig}). We validated it on real-world datasets.
The key contributions of this paper are outlined below: 
\begin{itemize}
    \item Exploration of using commonsense priors, particularly object size and orientation, and encoding them as prior factors for a graph-based SLAM system; % This approach represents a novel integration of intuitive human knowledge into robotic perception and mapping.
    \item A complete pipeline from embedding LLM-enhanced priors, extracting object-level features, performing data association, to solving incremental optimization;
    \item Evaluation using two real-world datasets: TUM RGB-D and 3RScan, and test of our SLAM method in real-time. % These evaluations were aimed at demonstrating the practical applicability and effectiveness of our proposed techniques in diverse and realistic settings.
\end{itemize}

\section{Related Work}
\label{sect:related}
\subsection{Object-level semantic SLAM}
%add orbslam and rtab-slam?
Semantic SLAM has evolved from treating semantic information as an ``add-on" to metric maps to integrating object constraints in trajectory optimization. While some approaches require dense point clouds \cite{hosseinzadeh2019real} or CAD models \cite{salas2013slam++}, generalized geometric representations have emerged for simplicity. The field predominantly uses two forms: quadric forms (ellipsoids) as in QuadricSLAM \cite{nicholson2018quadricslam} and cube forms as in CubeSLAM \cite{yang2019cubeslam}. CubeSLAM provides stronger constraints with rectangular 3D bounding boxes but struggles with irregularly shaped objects, while QuadricSLAM is more versatile for general object-level semantic SLAM.

Several works have built upon the QuadricSLAM approach \cite{wu2020eao, wu2023object, liao2022so, wang2024voom}, each with different emphases. The original QuadricSLAM required human-labeled data associations and solved through batch optimization. Qian et al. \cite{qian2021semantic} employed dense object keypoint features for automatic data association but did not focus on improving mapping quality. SO-SLAM \cite{liao2022so} emphasized first-frame initialization of objects and demonstrated the utility of spatial constraints. They evaluated their mapping techniques for initialization and used bundle adjustment with ground-truth data association for entire scenes. In contrast, our approach is tailored for real-time full-SLAM processing with online data association. EAO-SLAM \cite{wu2020eao} represents a similar work that adopts the CubeSLAM approach with line segmentation as joint localization methods, transforming cubes into quadric representations to map specific object classes. Meanwhile, EAO-SLAM also resolves data association via statistics test on dense visual features. More recently, VOOM \cite{wang2024voom} improved upon OA-SLAM \cite{zins2022oa} by employing a hierarchical coarse-to-fine approach that combines dual quadrics for object representation with enhanced feature point data association, demonstrating superior localization performance. Since both EAO-SLAM and VOOM optimize mapping with online data association, we compare our model with these systems in our experimental evaluation.

% \subsection{Object association}
% In SLAM processes, accurate data association is crucial for correctly linking sensor observations with objects or landmarks. When conducting online SLAM, landmarks are identified based on specific criteria rather than prior knowledge. Among various frame-to-frame data association algorithms, the Simple Online and Realtime Tracking (SORT) algorithm \cite{sort8296962} stands out for its simplicity, relying primarily on bounding box information and assumptions about object velocity. Several data association algorithms have been proposed using a probabilistic approach~\cite{doherty2020probabilistic,bowman2017probabilistic}, but these methods performed experiments only on a limited set of objects, such as chairs and doors, or cars in the outdoor dataset. Also, they only used and inferred the location of objects, rather than the full pose. Qian et al. \cite{qian2021semantic} uses the full 9 DOF quadric format, with dense object key point features for object association. However, they did not focus on the incremental object mapping, but rather used bundle adjustment like QuadricSLAM did. EAO-SLAM also introduced project object center position into 3D space and uses an ensemble method for data association. We use only bounding box information for short-term and long-term frame-to-frame associations, and show that the method is effective in the experiment. 

\subsection{Semantic methods with prior knowledge}
% Knowledge-based semantic methods
In robotics, prior knowledge is helpful across various disciplines\cite{fang2017object,pal2021learning}. Within the specific context of object-based SLAM, a follow-up work from QuadricSLAM authors attempted to use orientation \cite{jablonsky2018orientation} and then plane structures \cite{hosseinzadeh2019structure}  as constraints. 
% Here, researchers distinguish between vertical and horizontal orientations for common objects. 
Similarly, SO-SLAM \cite{liao2022so} leverages object-plane contact, object relative scales, and symmetry properties, showing an improved performance in initializing objects in SLAM. The key idea of our proposed method is to use prior factors not only for the initialization but throughout the optimization process because the constraints are useful in the incremental setting, especially when using geometric features in online data association. 

As LLMs have grown in capability and prominence, the robotics research community has increasingly leveraged their potential. Recent work has applied LLMs to various domains, including indoor object navigation\cite{shah2023navigation}, functional object search\cite{ge2024commonsense}, and tasks planning\cite{zhao2023large}. Studies\cite{yang2024thinking} have demonstrated that LLMs encode significant commonsense knowledge about physical objects and their properties. To the best of our knowledge, our work presents the first attempt to incorporate generalized object knowledge from LLMs directly into SLAM systems, bridging the gap between high-level semantic representations and geometric mapping.

\input{figure_latex/pipeline}

\section{Methodology}
\label{sect:method}
% Problem formulation
% We formulate the object SLAM problem with 
% TODO rephrase, mapping or slam?
% The two major problems we considered are the object association problem and the object mapping problem. 
% In this section, we propose methods to solve both of them incrementally and simultaneously with the sparse object-level features only.
% The association problem (long term/short term)
% TODO: only state the mapping problem? take association as a sub-problem to mention later

%need one section to introduce overall problem, what to achieve, etc. 

% We constructed a full object SLAM system to build an object-level map and estimate the agent's trajectory given visual perception.
Fig. \ref{fig:pipeline} presents an overview of our object SLAM system.
% Given a sequence of input RGB-D images, the visual odometry (VO) module provides odometry measurement by tracking visual features.
The odometry is estimated by tracking visual features from an RGB-D camera.
% Note that the object-level SLAM module is independent of the source of odometry measurements, and potentially we can utilize any other odometry sources such as an encoder or inertial measurement unit(IMU). However, image depth information is one type of prior factor in the semantic SLAM module. The depth features are also exploited when initializing object landmarks. Thus, it provides us with a compact implementation by directly using VO with RGB-D images.
% Since we require semantic visual features for the object SLAM module, we chose to use visual odometry (VO) for odometry estimation, while potentially any other odometry sources can be utilized such as an encoder or inertial measurement unit (IMU). 
% The object detection module provides 2D bounding box measurement with object semantic labels. 
Meanwhile, our system detects the per-frame 2D object features and performs inter-frame object tracking (short-term), and map-to-image object association (long-term).
The object SLAM module, utilizing a factor graph approach~\cite{factorgraph910572}, takes the estimated odometry and the associated object observations as input.
LLM is leveraged to generate commonsense priors for objects based on the vocabulary of the object detector, where the priors are modeled as unary factors in the factor graph (Fig. \ref{fig:factor-graph}). 
Finally, an incremental optimizer, such as iSAM2 \cite{kaess2012isam2}, updates the object map and the odometry.

% In this section, we first introduce the mathematical setup of the semantic SLAM problem. Based on the formulated SLAM problem, we propose our contextual prior factors and the data association method. We also present a unified method to construct a decent initial guess of the landmarks.

\subsection{Object SLAM problem}
\label{subsect:problem}
% probability break down, MAP probability and opt objective function
The object SLAM problem is set up with a sequence of camera poses $\mathcal{X} = \{\bm{x}_i \in SE(3) |i \in \{1, \cdots, M\}\}$ and the object landmarks, %represented by quadrics 
$\mathcal{Q} =\{\bm{q}_j \in \mathbb{R}^9 |j \in \{1, \cdots, N\}\}$. 
The odometry measurements, $\mathcal{U} = \{\bm{u}_i \in SE(3) | i \in \{1, \cdots, M-1\}\}$, represent the estimated pose transformation between two consecutive camera poses, $\mathbf{x}_{i+1}=f\left(\mathbf{x}_i, \mathbf{u}_i\right)+\mathbf{w}_i$. % With the VO measurements, 
The motion model $f(\cdot)$ is simply $f\left(\mathbf{x}_i, \mathbf{u}_i\right) = \mathbf{x}_i \oplus \mathbf{u}_i$, where the $\oplus$ operator denotes the summation operation over SE(3). $\mathbf{w}_i$ is the Gaussian white noise with covariance $\Sigma_f$, i.e., $\mathbf{w}_i \sim \mathcal{N}(\bm{0}, \Sigma_f)$.

For the object landmarks, we follow the formulation as in QuadricSLAM \cite{nicholson2018quadricslam}, where ellipsoids represent all landmarks. $\bm{q}_j$ characterizes the 9 degrees of freedom (DoFs) for a quadric, i.e., 
the rotation angles $\bm{\theta} = (\theta_x, \theta_y, \theta_z)^T$, the quadric centroid translation $\bm{t} = (t_x, t_y, t_z)^T$, and the lengths of 3 semi-axes $\bm{s} = (s_x, s_y, s_z)^T$, %. $\bm{q}$ is obtained by stacking the parameters together as
where $\bm{q} = (\bm{\theta}^T, \bm{t}^T, \bm{s}^T)^T$.
% Note that a quadric $\bm{q}$ can be defined by a $4\times4$ symmetric matrix $\bm{Q}$ as well, %, where its dual form $\bm{Q}^*$ is convenient for expressing the projected dual conic 
% (projecting a quadric to an image plane leads to a conic).
% on an image plane.
The object landmarks are observed as on-image bounding boxes using visual object detectors such as the YOLO\cite{Jocher_Ultralytics_YOLO_2023}.
The object observations are denoted as $\mathcal{B} = \{\bm{b}_{ik} \in \mathbb{R}^4 |k \in \{1, \cdots, L_i\}\}$ 
where $L_i$ is the number of observed objects in the $i$-th frame, $\bm{b}_{ik}$ represents the $k$-th bounding box in the $i$-th frame. %, and $\bm{q}_j$ is the $j$-th landmark from the historical observations.
Each bounding box is represented as 
$\bm{b}=(x_{\text{min}}, y_{\text{min}}, x_{\text{max}}, y_{\text{max}})^T$. %, which depicts the rectangular area for object observation in the image pixel unit.
Given $\mathbf{x}_i$ and $\mathbf{q}_j$, a predicted bounding box $\hat{\bm{b}}$ is obtained by projecting $\mathbf{q}_j$ onto the image plane. The observation model is denoted as $\hat{\bm{b}}_{ij} = \boldsymbol{\beta}\left(\mathbf{x}_i, \mathbf{q}_j\right)$, where implementation is detailed in \cite{nicholson2018quadricslam}.

We denote the prior and initial measurements of each quadric as $\bm{\alpha}_j$, and $\mathcal{A} = \{\bm{\alpha}_j | j \in \{1, \cdots, N\}\}$. %, where we rigorously define $\bm{\alpha}_j$s later in this section. % Markov assumption
The probability distribution over all camera poses and quadric landmarks can be factorized as:

\begin{equation}
\begin{aligned}
P(\mathcal{X}, \mathcal{Q} \mid \mathcal{U}, \mathcal{B}, \mathcal{A}) \propto & \prod_i P\left(\mathbf{x}_{i+1} \mid \mathbf{x}_i, \mathbf{u}_i\right)  \\
\cdot & \prod_{i j} P\left(\mathbf{q}_j \mid \mathbf{x}_i, \mathbf{b}_{i j}\right) \cdot \prod_{j} P \left( \bm{q}_j | \bm{\alpha}_j \right)
\end{aligned}
\label{eq:cond-prob}
\end{equation}

% By Bayes' theorem, $P(\bm{q}_j|\bm{x}_i, \bm{b}_{ij}) \propto$ $ P(\bm{b}_{ij} | \bm{q}_j, \bm{x}_i) \cdot P(\bm{q}_j | \bm{x}_i)$,
% where $P(\bm{q}_j | \bm{x}_i) = P(\bm{q}_j)$
% since the camera trajectory and landmark locations are independent.
We can then formulate the maximum a posteriori (MAP) problem by maximizing the joint probability in (\ref{eq:cond-prob}):

\begin{equation}
\begin{aligned}
\mathcal{X}^*, \mathcal{Q}^*=\underset{\mathcal{X}, \mathcal{Q}}{\operatorname{argmin}} & -\log P(\mathcal{X}, \mathcal{Q} \mid \mathcal{U}, \mathcal{B}, \mathcal{A}) \\
=\underset{\mathcal{X}, \mathcal{Q}}{\operatorname{argmin}} & \sum_i\left\|f\left(\mathbf{x}_i, \mathbf{u}_i\right) \ominus \mathbf{x}_{i+1}\right\|_{\Sigma_f}^2 \\
& +\sum_{i j}\left\|\mathbf{b}_{i j}-\boldsymbol{\beta}\left(\mathbf{x}_i, \mathbf{q}_j\right)\right\|_{\Lambda_{i j}}^2 \\
& +\sum_{j} \left\| g(\bm{q}_j) - \bm{\alpha}_j \right\|_{\Sigma^{(j)}_{\alpha}}^2
\end{aligned}
\label{eq:opt-question}
\end{equation}
where the $\ominus$ operator denotes the difference operation in SE(3), $\left\| \mathbf{a} - \mathbf{b} \right\|_{\Sigma}$ is the Mahalanobis distance with covariance matrix $\Sigma$, and $g(\cdot)$ represents a general transform function on the quadric $\bm{q}_j$ such that $g(\bm{q}_j)$ can be directly compared with the prior estimates.
% With the factorized conditional probabilities (\ref{eq:cond-prob}) and MAP formulation of the objective function (\ref{eq:opt-question})

\subsection{Factor graph and LLM-enhanced priors}
\label{subsect:prior}

With the MAP problem setup, we can utilize factor graphs to characterize the object-level SLAM efficiently.
Fig. \ref{fig:factor-graph} shows a toy example of the factor graph structure with our commonsense prior factors. 
We introduce the size and orientation priors, deriving from commonsense knowledge using LLM. Besides, we designed a unary factor with initial object centroid measurement, drawing from a single-frame RGB-D observation of objects, to facilitate fewer-frame object SLAM optimization.

\input{figure_latex/factor}

\subsubsection{Size prior}
\label{subsubsect:size-prior}

% We use prior semantic information of each object to inquire about the common-sense knowledge of the object size. 
Human language mostly describes the size of an object using length, width, and height, while axis lengths most easily describe the size of ellipsoids. 
We use LLM to generate the length, width, and height for a list of object categories. 
However, an object can have different rotations, and a fixed correspondence between axis lengths and length, width, and height is not guaranteed.
% (e.g., a book can be put vertically on a bookshelf but can also lie on a table). 
Therefore, we re-ordered the landmark size measurements to determine the dissimilarity between the axis length estimation and the size prior. 
The re-ordering operation is denoted by a permutation matrix $P_s$, such that $\bm{\hat{s}} = P_s \bm{s}$, where $\bm{\hat{s}}$ contains all entries in $\bm{s}$ in ascending order.
% Denote the ordering operation on a 3-dimensional size vector $\operatorname{ord}(\bm{x}) = (\operatorname{max}(\bm{x}), \operatorname{median}(\bm{x}), \operatorname{min}(\bm{x}))^T, \forall \bm{x} \in \mathbb{R}^3$. 
For a given commonsense size of an object $\bm{s}_r$ in the format of length-width-height, the respective prior estimate of size is
$\bm{s}_{\alpha} = P_s \bm{s}_r / 2$.
% \begin{equation}
%     \bm{s}_{\alpha} = \operatorname{ord}(\bm{s}_r) / 2
% \end{equation}
We divide $\bm{s}_r$ by 2 to be consistent with the representations of the semi-axis length for a quadric.

In practice, we found that the effectiveness of the size prior is highly dependent on the accuracy of the object label, which is provided by the object detector. Hence, we combined $\bm{s}_{\alpha}$ with the initial size of the quadric $\bm{s}_o$, which is derived from the initial bounding box dimensions and the depth estimation within the bounding box. We formulated the prior estimate as $\hat{\bm{s}}_{\alpha} = p \cdot \bm{s}_{\alpha} + (1-p)\bm{s}_o$, where $p \in [0, 1]$ is the confidence score for the object observation given by the object detector. Note that state-of-the-art object detection algorithms can have high confidence score even for partially observed objects, and thus our size prior is necessary to correct such occluded observations.

\subsubsection{Orientation prior}
The fundamental rule for orientation prior factors incorporates that the rotation of real-world objects is aligned with gravity. Moreover, based on the affordance of objects, an object can have (1) vertical, (2) horizontal, or (3) uncertain orientation. Vertical objects have a height greater than their length and width, i.e., the object size along the z-axis is larger than their x- and y-axes. For instance, bottles and chairs are normally upright because of their designed functions and usually have height as the longest dimension. On the other hand, horizontal objects, such as keyboards and plates, have a height significantly smaller than their length and width. Under the uncertain orientation category, objects can either be common to manipulate into various rotations (e.g., books and teddy bears) or have all three dimensions in similar lengths (e.g., balls and bowls). The rotation of the ellipsoid representation of the latter ones is not interesting for optimization since the shape of the ellipsoid will be very close to a sphere, and we simply follow the base rule of the gravity constraint.

For vertical objects, we align the longest semi-axis of the quadric with the z-axis of the world frame, $\bm{w}_z = \left[0, 0, 1\right]^T$, where the semi-axis lengths are obtained from initial observation $\bm{s}_o$. The longest axis of the horizontal objects is projected to the world x-y plane. 

For uncertain objects, we perform gravity alignment with minimum projection error.
% The quadric rotation is aligned with the z-axis of the world frame, $\bm{w}_z = \left[0, 0, 1\right]^T$. %, and we use the aligned rotation as the prior orientation $\bm{\theta}_{\alpha}$. 
Denote the initial quadric orientation using rotation matrix $R_q = \left[\bm{r}_x, \bm{r}_y, \bm{r}_z\right]$, and the gravity-aligned quadric rotation $\hat{R}_q = \left[\hat{\bm{r}}_x, \hat{\bm{r}}_y, \bm{w}_z\right]$. 
We set 
$\hat{\bm{r}} = \operatorname{argmin}_{ \bm{r} \in (\bm{r}_x, \bm{r}_y, \bm{r}_z)} |\bm{r}^T \bm{w}_z|$
, and hence $\hat{\bm{r}}$ is the initial rotation axis closest to the world x-y plane.
The orthogonal projection of $\hat{\bm{r}}$ onto the world x-y plane gives the minimum projection error in $L_2$-norm while revising the quadric orientation to be aligned with gravity. We estimate $\hat{\bm{r}}_y$ by a Gram-Schmidt step, i.e.,
% the projected vector is normalized to unit length to represent an axis. This operation is defined as: % TODO should we just use  Gram-Schmidt to describe it? (but it has different meaning than gram-schmidt)
    $\hat{\bm{r}}_y = \frac{\hat{\bm{r}} - \left(\hat{\bm{r}}^T \bm{w}_z\right) \bm{w}_z}{\left\|\hat{\bm{r}} - \left(\hat{\bm{r}}^T \bm{w}_z\right) \bm{w}_z \right\|_2}$,
% wy = ax - np.dot(ax, wz)*wz
where $\hat{\bm{r}}_y$ defines the gravity-aligned y-axis of the quadric. 
% Finally, 
$\hat{\bm{r}}_x$ can be recovered by $\hat{\bm{r}}_x = \hat{\bm{r}}_y \times \bm{w}_z$. 
% The prior estimate of a quadric's rotation can be obtained by $\bm{\theta}_{\alpha} = \left(log(\hat{R}_q)\right)^{\vee}$, where $log(\cdot)$ is the matrix logarithm and $(\cdot)^{\vee}$ stands for the Vee map operator.
% Note that it is equivalent to set the projected axis as $\hat{\bm{r}}_x$ and find $\hat{\bm{r}}_y$ by $\hat{\bm{r}}_y = \bm{w}_z \times \hat{\bm{r}}_x$ and change the order of the quadric semi-axis lengths $\bm{s}$ correspondingly.

\subsubsection{Centroid factor}
For the centroid factor, we encode the initial guess of the object centroid $\bm{t}_{\alpha}$ as a unary factor. 
We utilize the depth image to obtain the initial guess of the object's centroid location. %Section \ref{subsect:init} provides a detailed description of the initialization method. 
The centroid factor is used to constrain the position of a quadric, especially in the case of occlusion. 
% We use the depth information as prior factors to assist the inference since it would be hard to get an accurate landmark estimate at the beginning when the camera does not have enough observations. 

\subsubsection{Cost function and variance settings}
We denote the cost term for prior factor and initial measurement in (\ref{eq:opt-question}) as
$e_{\alpha}^{(j)} := \left\| g(\bm{q}_j) - \bm{\alpha}_j \right\|_{\Sigma^{(j)}_{\alpha}}^2$.
For any landmark $\bm{q}$:
\begin{equation}
\begin{aligned}
        e_{\alpha}
        &= \left\| \bm{\theta} - \bm{\theta}_{\alpha} \right\|_{\Sigma_{\theta}}^2 + \left\| \bm{t} - \bm{t}_{\alpha} \right\|_{\Sigma_t}^2 + \left\| P_s \bm{s} - \bm{s}_{\alpha} \right\|_{\Sigma_s}^2 
        % \\
        % &=: \left\| g(\bm{q}) - \bm{\alpha} \right\|_{\Sigma_{\alpha}}^2
\end{aligned}
\label{eq:prior-cost}
\end{equation}
where the joint covariance matrix
$\Sigma_{\alpha} = \text{diag}(\Sigma_{\theta}, \Sigma_t, \Sigma_s)$, the transform function $g(\bm{q})=(\bm{\theta}^T, \bm{t}^T, P_s \bm{s}^T)^T$, and  $\bm{\alpha} = (\bm{\theta}_{\alpha}^T, \bm{t}_{\alpha}^T, \bm{s}_{\alpha}^T)^T$.

% covariance proportional to box size / quadric size 
% As we can see from (\ref{eq:prior-cost}), 
% The cost term $e_{\alpha}$ is affected by the covariance matrix $\Sigma_{\alpha}$. %, which includes some hyperparameters for us to tune.
In terms of covariance setting, we set $\Sigma_{\theta}=\operatorname{diag}(k_x^2, k_y^2, k_z^2)$ with $k_x, k_y \in (0, \frac{\pi}{16})$ and $k_z \in (\frac{\pi}{2}, \pi)$. 
% We set a relatively large prior noise variance to the orientation factor, so that the incremental solver can easily recover from a biased prior. 
% However, the noise level should not be too large such that it cannot provide enough constraint for the incremental solver to work when the observations are very partial at the beginning.
We also set $\Sigma_t$ and $\Sigma_s$ such that the standard deviation is proportional to the inquired object size, i.e., $\Sigma_t = k_t^2 \cdot \operatorname{diag}(\bm{s}_{\alpha})^2$ and $\Sigma_s = k_s^2 \cdot \operatorname{diag}(\bm{s}_{\alpha})^2$. $k_t$ and $k_s$ are set in the range of $(0.5, 2)$.
% TODO explanation of the ranges?

\subsection{Object association}
\label{subsect:association}

The object-level data association is implemented by coordinating on-image object tracking (e.g., SORT \cite{sort8296962} and its variations \cite{deepsort8296962, aharon2022botsort}) and the matching between landmarks and bounding boxes for long-term association. 
% Numerous existing research proposed on-image tracking algorithms, such as SORT \cite{sort8296962} and some of its variations \cite{deepsort8296962, aharon2022botsort}, where a tracking ID is assigned to each 2D bounding box detection, and detections with the same tracking ID indicate that they are associated to the same object.
The long-term data association task is modeled as a linear sum assignment problem (LSAP).
% short-time tracking - sort \\
% long-time tracking - IOU + 
Specifically, LSAP finds the bijection between all $\bm{b}_i$ and $\bm{q}$, such that the summation of weights $w_{kj}^{(i)}:=w(\bm{b}_{ik}, \bm{q}_j)$ over all pairs of $\bm{b}_{ik}$ and $\bm{q}_j$ can be maximized.

Define a weight matrix $W^{(i)}=(w_{kj}^{(i)})$ for data association: 
\begin{equation}
    w_{kj}^{(i)} = \operatorname{IoU}(\bm{b}_{ik}, \beta (\bm{x}_i, \bm{q}_j)) + w_s \cdot \psi_{k j}^{(i)} + w_d \cdot \frac{1}{||\bm{t}_j - \hat{\bm{t}}_{i k}||_2}
    \label{eq:asso-weight}
\end{equation}
where $\operatorname{IoU}$ represents Intersection-over-Union, $w_s$ is a constant semantic weight, %TODO (to resolve occlusion on 2d projection)
and $\psi_{k j}^{(i)}$ is defined as:
\begin{equation}
    \psi_{k j}^{(i)} = \begin{cases} 1 & \text{if } ||\bm{v}^b_{ik}-\bm{v}^q_{j}||_2 \leq s_{th} \\ 0 & \text{otherwise} \end{cases}
\end{equation}
where $\bm{v}$ is the GloVe word embedding \cite{pennington2014glove}
for the class label of a bounding-box detection $\bm{b}_{ik}$ or a landmark $\bm{q}_j$, and $s_{th}$ is a semantic threshold. In this case, a higher association weight is assigned to the detection and landmark with similar class labels.
% explain the distance term w_d = 1/200
$\hat{\bm{t}}_{i k}$ represents the reprojected centroid from a bounding box observation to a 3D point in the world frame. The reprojection procedure adopts the depth image information and uses the same method as obtaining the centroid factor estimation. $w_d$ is a hyperparameter for scaling.
The last term in (\ref{eq:asso-weight}), which is inversely proportional to the distance between an ellipsoid centroid and a reprojected centroid, identifies data association when an object has a relatively small size and is hard to track using IoU.
% TODO rephrase this 
Combining the weights of IoU, semantic matching, and distance, we have the weight function for LSAP as $w(\bm{b}_{ik}, \bm{q}_j) = w_{kj}^{(i)}$, where solving the LSAP yeilds the long-term object-level association.
% cost matrix (change max to a min problem) $\mathcal{C}^{(i)} = (-w_{kj}^{(i)})$

% \subsection{Object initialization}
% \label{subsect:init}
 
% We introduce the following scheme to initialize a quadric from a single-frame observation utilizing the depth image:
% \begin{itemize}
%     \item For the quadric centroid, the x and y coordinates are initialized to be the bounding box center, and the z coordinate is set to be the average depth in the bounding box, where the depth data is filtered so only data within the interquartile range (IQR) is used for calculation;%. The centroid translation $\bm{t}$ can be obtained by transforming the estimated centroid to the world frame;
%     \item Rotation is first assigned using the camera's rotation (i.e., semi-axes of the quadric are aligned with the bounding box) and then rectified to be gravity-aligned; % using the same method as computing the orientation factor;
%     \item The quadric semi-axes $s_x$ and $s_y$ are the lengths in meters from the bounding box margins to its center. $s_z$ is calculated by using $_{\{C\}} t_z$ (the translation in the z-direction of the camera frame) to subtract the minimum depth inside the bounding box;
%     \item  The bounding boxes at image margins are discarded for initialization to avoid partial observations. 
%     % However, if objects appear in the center of an image but are detected with occlusions  
%     % Such cases are discussed in more detail in
%     % see Section \ref{sect:exp};
% \end{itemize}

\subsection{Generating Object Priors using Large Language Models}
\label{subsect:llm}
We use LLMs to generate commonsense object priors, specifically size dimensions and orientation information, to enhance object-level SLAM performance. Our implementation used Claude 3.7 Sonnet. The methodology is also generalizable to other LLM architectures with sufficient semantic knowledge.

LLMs encode substantial commonsense knowledge about the physical world through their training on vast text corpora. We exploit this capability to extract structured information about everyday objects without requiring manual annotation or specialized databases. The prompt is formulated to elicit commonsense physical knowledge while constraining the response format for systematic processing:
\begin{quote}
\small\texttt{Provide commonsense knowledge about the physical dimensions and orientation of the following objects. Generate a CSV with columns (object, length, width, height, orientation) where dimensions are in meters, orientation is coded as 0=vertical, 1=horizontal, 2=uncertain. Preserve the exact object names from the input list and maintain the same number of objects.}
\end{quote}
 %  We applied this methodology to extract priors for objects from the COCO dataset (~80 object categories) and the 3RScan dataset (~500 object categories).
The generated priors are integrated directly into our object-level SLAM pipeline as prior estimations. This information serves as soft constraints during optimization, helping to improve initial object pose estimates and constrain solutions when observations are ambiguous.

% The quality of LLM-generated object priors is evaluated using Mean Absolute Error (MAE) against human-labeled ground truth from the 3RScan and TUM datasets, with detailed results presented in the Experiments section.

% \subsection{Failure-proof mechanism for mismatching}

\section{Experiment}
\label{sect:exp}
We evaluated the performance on two real-world RGB-D datasets for indoor scenes. 
Our full object SLAM system was tested on the TUM RGB-D dataset \cite{tum2012rgbd-dataset}. 
We validated the effectiveness of our LLM-enhanced priors for online updates on 3RScan~\cite{Wald2019RIO}. 
% Furthermore, we performed tests in the real world and analyzed error and failure cases. 
Furthermore, we conducted an ablation study to discuss how each prior affects the mapping results.

\subsection{TUM RGB-D}
\label{subsect:tum-rgbd}

\input{table/tum-map}
% \vspace{-\baselineskip}

\input{table/tum-association}
\vspace{-12mm}

\textbf{Dataset}
TUM RGB-D dataset is a well-known indoor SLAM benchmark with the ground-truth trajectory provided for test and evaluation.
We tested three sequences in the TUM RGB-D dataset under the ``Handheld SLAM" category: \textit{freiburg1\textunderscore desk} (fr1), \textit{freiburg2\textunderscore desk} (fr2), and \textit{freiburg3\textunderscore long\textunderscore office\textunderscore household} (fr3), which are office scenes with cluttered objects. 

\textbf{Experiment setting}
We adopted ORB-SLAM3 \cite{Campos_2021orb3} to provide the visual odometry, YOLO-v8 \cite{Jocher_Ultralytics_YOLO_2023} as the object detector, and the SORT algorithm for short-time tracking. Since the TUM RGB-D dataset does not provide the ground-truth object location and pose, we manually labeled the objects that could be detected by the YOLO-v8 detector, where labelCloud \cite{sager2021labelcloud} is the 3D object annotation tool we used here.

We compared our SLAM algorithm with EAO-SLAM and VOOM. Same object detection model was used to ensure a fair comparison. We selected YOLOv8x-seg model since VOOM requires an object detector with segmentation. We also tested our object SLAM with the open-vocabulary YOLO-World detector \cite{Cheng2024YOLOWorld} by feeding a list of common indoor objects as the vocabulary, demonstrating the strength of our LLM-enhanced priors when accurate object category labels are available. 

\textbf{Evaluation metrics}
% Since most of the existing real-world datasets for visual SLAM experiments are not specifically designed for object SLAM, 
We found that there is an absence of evaluation metrics for semantic mapping in previous research~\cite{nicholson2018quadricslam, wu2020eao}, while we conducted quantitative analysis using our annotated 3D object bounding boxes.
We evaluate the mapping quality by comparing the size and centroid location of the ellipsoid landmarks with the labeled object instances.
The centroid error of a landmark is calculated by the Euclidean distance between the estimated quadric centroid and the corresponding centroid of the labeled 3D bounding box.
Similarly, we define the size error as the L$_2$-norm of difference in quadric size.
We also calculated 3D bounding-box IoU using \cite{ravi2020pytorch3d} for oriented 3D boxes. We compared how the 3D IoU error performance varies over time, where IoU error is simply $1-\text{IoU}$.
The mapping results were evaluated for the estimated objects that are correctly associated.
In all of the tables that involve object mapping evaluation, the above three metrics are denoted as ``Centroid", ``Size", and ``3D IoU" respectively.
%, object centroid error $\epsilon_d$, and object size error $\epsilon_s$, where the latter two metrics are the same as defined in Section \ref{subset:3rscan}. 

% In addition, we evaluated the effectiveness of the data-association algorithm by counting the number of object instances in the final maps. 
Although the main contribution here is not about improving object-level data association, it is important to notice that data association and mapping are tightly coupled. Data association can be compromised if object mapping from the first few observations is unsatisfactory, while poor data association also leads to fewer observations per object, which can severely affect the final object mapping quantity.
Different from the data-association evaluation in most of the previous research, which only counts the total number of objects that are mapped \cite{wu2020eao, qian2021semantic}, we conducted a more comprehensive evaluation that differentiates the True Positive (TP), False Positive (FP), and False Negative (FN) -- metrics that are often used in the 2D data-association research. TP ensures a one-to-one association between the estimates and ground-truth data. FP refers to the estimated objects that are not in the ground-truth data, and FN counts the number of ground-truth objects which are absent in the estimated map.  

We compare the Absolute Trajectory Error (ATE) of the estimated camera trajectory against the provided ground-truth poses. Since we adopt ORB-SLAM3 as the source of visual odometry, the ATE from ORB-SLAM3 (represented by ``ORB3") is also calculated as a reference. We also assess the real-time performance of our object SLAM system by the average frame rate, and the breakdown timing on odometry, detection, association, and incremental optimization.

\input{figure_latex/tum_incremental}

\input{figure_latex/tum_iou}
% \vspace{-1mm}

\textbf{Results} 
Table \ref{table:tum-map} demonstrates our object mapping results compared to EAO-SLAM and VOOM. The entries ``EAO", ``VOOM", and ``Ours" refer to using EAO-SLAM, VOOM, and our object SLAM method with YOLOv8x-seg detection model correspondingly, and ``Ours+W" used our SLAM system with YOLO-World detector. With YOLO-World detector, our SLAM system outperforms other baselines on mapping results across all scenes and achieved the lowest ATE on \textit{fr2} and \textit{fr3} sequences. Since \textit{fr1} has few object observations, EAO-SLAM has the advantage of using line detection and dense visual feature. Our SLAM is purely based on ORB-SLAM3 odometry and sparse object-level feature, yet we achieved a similar ATE performance compared with ORB-SLAM3. 

We underscored the second best performance in Table \ref{table:tum-map} so that we can compare among the SLAM performance using YOLOv8x-seg detection. Under the same detector setting, our SLAM still exceeds the baselines on most of the metrics and sequences. We demonstrate an average of 36.8\% increase in 3D IoU compared to VOOM and 78.2\% compared to EAO-SLAM. VOOM achieves 7.9\% less in size error, but sacrifices in all other metrics by a large margin.

Table \ref{table:tum-association} shows the data association results. ``Total" refers to the total count of labeled ground-truth (GT) objects. E, V, O, and W are short for EAO-SLAM, VOOM, Ours, and ``Ours+W". EAO-SLAM tends to aggressively associate the object observations and merge multiple objects into one, consequently having a lot of FN results. Conversely, VOOM has a poor data associator. Although VOOM covers the most GT objects, it generates abundant FP objects as well. Fig. \ref{fig:tum} is a direct comparison between VOOM's and our final object map. Our object map is cleaner and mostly reflects the ground-truth objects. 
We calculated the average IoU error among 10 selected objects that have the most correctly associated 2D features for both VOOM and our SLAM. The IoU error comparison is demonstrated in Fig. \ref{fig:tum-iou}. Our method has better initialization results, and the IoU error decreases faster. With extra constraints from prior factors, our SLAM also exhibits a more consistent descent in IoU error than the trend of VOOM, where the latter method is more vulnerable to poor bounding box observations. During the first 10 seconds from the initial observation of a certain object, the 3D IoU error of our SLAM decreases by 17.1 \% and remains similar afterward. 

We evaluated the runtime performance of our algorithm on a Laptop with an Intel Core i9 CPU and an Nvidia GeForce GTX 3080 GPU (GPU is only required for YOLO-v8 detection. All the other calculations are on CPU). On average, visual odometry requires 18 ms per frame, the object detector takes 24 ms, data association needs 7 ms, and factor graph optimization costs 16.5 ms. The processing speed decreases as more observations are added to the factor graph, dropping from 20 Hz to 2.5 Hz. The overall speed of the online SLAM system averages around 10 Hz on \textit{fr3} sequence.

\subsection{3RScan}
\label{subset:3rscan}

\textbf{Dataset} The 3RScan dataset consists of 1482 scans across 478 different scenes. Each scene was recorded by a handheld Google Tango cellphone, yielding sequences of RGB-D images with calibrated camera poses. 3RScan includes 3D reconstruction of object segmentation, allowing quantitative evaluation of our mapping performance.

\textbf{Experiment setting} On the 3RScan dataset, we experimented on 57 scans, which were selected considering feature quantity and diversity.
% (i.e., we want to make sure there are enough object features detected and the 2D features belong to various types of objects). 
% The experiment was repeated 20 times to estimate an average behavior.
% of our online update algorithm.
The 3RScan dataset is designed for 3D scene reconstruction and assumes camera odometry has been resolved. The consecutive frames are sampled sparsely, and it is challenging to track the visual features to extract the trajectory or maintain object-level association. We evaluated the mapping results from incremental update of our SLAM algorithm using ground-truth odometry and data association to demonstrate the effectiveness of using priors. 
% Our full SLAM system with data association was assessed on the TUM RGB-D dataset (Section \ref{subsect:tum-rgbd}).

We compared our full SLAM system with the setting under no prior factors and centroid measurements (referred as ``w/o priors" in Table \ref{table:3rscan}). %``Ours" refers to the algorithm that incorporates improved initialization and prior factors.
%, thereby facilitating the process of incremental optimization. 
While the focus is on improving real-time performance, we also compared our algorithm against the same ``w/o priors" setting using batch optimization.
% In batch optimization scenarios, since objects can be initiated from multiple views across all frames, we employ the Singular Value Decomposition (SVD) solutions for the initializing quadric landmarks in 3D space, so that improvements on single-frame initialization are redundant. 

\input{figure_latex/mapping_3RScan}
% \vspace{-8mm}

\textbf{Evaluation metrics} 
We use the same evaluation metrics for object mapping as in Section \ref{subsect:tum-rgbd}.

\input{table/3RScan}

\vspace{-8mm}

\textbf{Results} 
Table \ref{table:3rscan} presents our experimental results on the 3RScan dataset, demonstrating that our prior-assisted method outperforms no-prior system across all metrics. For online updates, our method reduces object centroid error by 59.6\% and object size error by 70.1\% compared to the no-prior case. Additionally, our approach achieves higher 3D IoU. Besides, our online method still shows superior performance when compared against no-prior method using batch optimization, with 41.9\% lower in centroid error and 52.0\% lower in size error. These results validate the effectiveness of our prior factors.%, particularly considering that our method is primarily designed for online updates yet outperforms even the batch optimization approach.

Fig. \ref{fig:3RScan} demonstrates the result of the final maps using online updates for qualitative analysis. Most of the estimated objects are consistent with the segmented instances. There are some objects on the ceilings in Fig. \ref{subfig:3rscan-bedroom2} and Fig. \ref{subfig:3rscan-living2} that were mapped with displacement compared to the bounding boxes from 3D segmentation. These instances are partially observable and only appear in very few camera frames, and thus it is difficult for the algorithm to initialize and optimize their positions accurately. 
% We chose ground truth association for the 3RScan dataset due to the varying quality of images captured by mobile phones, often characterized by low resolution and blurriness, which significantly impact the performance of deep-learning object detectors.

\subsection{Evaluation of LLM-Generated Object Priors}
To assess the quality of object priors generated by LLMs, we evaluated the dimensional and rotational accuracy across objects from the TUM RGB-D dataset (27 object types) and 3RScan (222 object types) by comparing with the human-labeled objects. 3RScan dataset provided object annotations, which we found thin objects like papers or notebooks often appear thicker than their actual dimensions for better visualization effects. We manually labeled the objects in the TUM RGB-D dataset on the dense 3D reconstructed scene for each sequence, where a few objects also appear larger than actual object dimensions because of the noise in reconstructed point clouds. 
% Ground-truth measurements were obtained from human annotations in the 3RScan and TUM datasets. 

We computed the Mean Relative Error (MRE) between LLM-predicted dimensions and ground truth measurements, finding an average error of 35.2\% and 30.8\% for 3RScan and TUM RGB-D, respectively. With further analysis by grouping in categories of 3RScan objects, we found that furniture has the lowest MRE of 24.5\%, while decorative and stationary objects have the highest MRE of 64.7\%. Additionally, 77.5\% of objects have a larger labeled volume than the LLM-predicted one, which validates our observation on the labeled objects tend to appear larger and suggests that the labeling inaccuracy partially contributes to the MRE as well. Nevertheless, mechanisms are designed for our LLM-enhanced priors to handle mismatches between LLM-generated sizes and the observations (Section \ref{subsubsect:size-prior}).
% This discrepancy is reasonable since the ground truth was labeled from point cloud reconstructions of the scene, where thin objects like papers or notebooks often appear thicker than their actual dimensions.

For orientation classification (vertical, horizontal, or uncertain), we used both strict and weighted metrics. Strict accuracy requiring exact matches reached 66.7\% (TUM) and 64.8\% (3RScan). Our weighted metric, assigning partial credit (0.5) when either ground truth or prediction is ``uncertain," yielded improved accuracies of 77.8\% (TUM) and 68.9\% (3RScan), indicating that many predictions are reasonable approximations despite not being exact matches.

\input{table/ablation}
\vspace{-8mm}

\subsection{Ablation study of prior factors}

We conducted an ablation study on the effectiveness of having only one of the prior factors in the object SLAM system. 
The test was conducted on \textit{fr2\textunderscore desk} and \textit{fr3\textunderscore office} sequences because these sequences contain the most number of objects among the TUM RGB-D sequences. YOLO-World detector is selected here to demonstrate the best usage of our prior factors.

Table \ref{table:ablation} presents the results of our ablation study using the same object mapping metrics as in Section \ref{subsect:tum-rgbd}. N, O, S, and F represent respectively the cases when \textit{neither} of the prior factors are deployed, \textit{orientation} prior is used but size prior is eliminated, \textit{size} prior is adopted but orientation prior is missing, and \textit{full} object SLAM system with both priors, which is under the same setting as Ours+W in Table \ref{table:tum-map}.

Table \ref{table:ablation} shows that orientation prior is effective for achieving higher 3D IoU, but may compromise the accuracy in object centroid and size, such as the results in \textit{fr2}. We conjecture that the optimizer attempts to fit quadrics into the bounding box observations with a certain orientation, and even if that causes vastly expanding the ellipsoid size along the camera depth direction. We also found that the object mapping results on \textit{fr3} are similar when one or both prior factors are missing. This is because the incremental optimization solver sometimes lacks enough constraints to perform online updates when there are not enough prior estimations, leading the final object maps to remain similar to their initialization results.
Hence, it is crucial to deploy both prior factors together for the best performance of our object SLAM system.

% \subsection{Real World Experiment and Error Analysis}
% We performed real-world experiments in the school laboratory area. The 

\section{Conclusion}
\label{sect:conclusion}
In conclusion, we proposed a real-time object SLAM framework incorporating LLM-inspired prior knowledge as constraints for online factor graph optimization. We demonstrated that our algorithm operates effectively in real-time and real-world scenarios. Using prior factors enhances object-level SLAM results, and analysis showed that LLM-generated priors align well with real-world data. 

In future work, we will also explore more in-depth the fusion of image detection methods and point cloud-based methods regarding different categories of objects. Also, data association can achieve greater robustness through scene-level matching in addition to matching at the object level. We also plan to integrate the spatial relationships into a scene graph to establish a more comprehensive understanding. We will ultimately integrate our SLAM with the object search method for the online object-goal navigation task. 


%%%%%%%%%%%%%%%%%%%%%%%%%%%%%%%%%%%%%%%%%%%%%%%%%%%%%%%%%%%%%%%%%%%%%%%%%%%%%%%%

%%%%%%%%%%%%%%%%%%%%%%%%%%%%%%%%%%%%%%%%%%%%%%%%%%%%%%%%%%%%%%%%%%%%%%%%%%%%%%%%
% \vspace{5}
\bibliographystyle{IEEEtran}
\small{
\bibliography{reference}
}
\vfill
\end{document}

%% file: figure_latex/pipeline.tex
\begin{figure*}[tbp]
    % \centering
    \centerline{\includegraphics[width=0.95\textwidth]{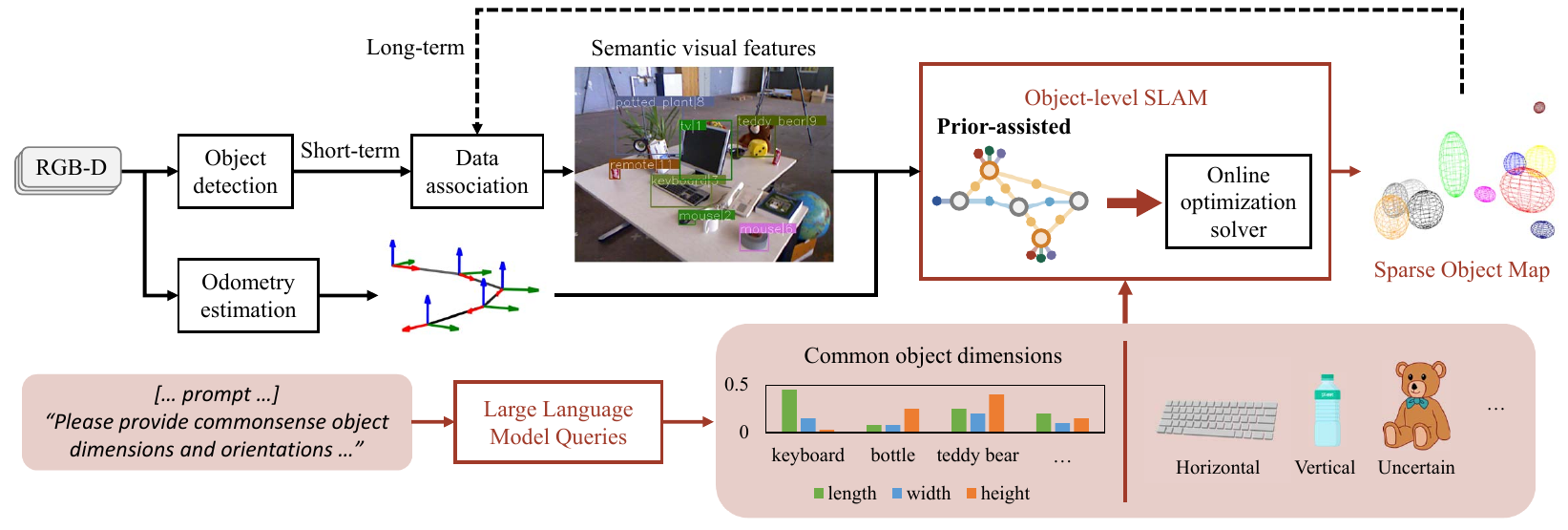}}
    \caption{The pipeline of our object SLAM method. We describe our SLAM system by showing the usage of LLM and the data flow from a sequence of image frames to the object-level map. Our main contributions are highlighted in red.}
    \label{fig:pipeline}
\end{figure*}

%% file: figure_latex/factor.tex
\begin{figure}[tb]
    \setlength{\belowcaptionskip}{-14pt} 
    % \centering
    \centerline{\includegraphics[width=0.4\textwidth]{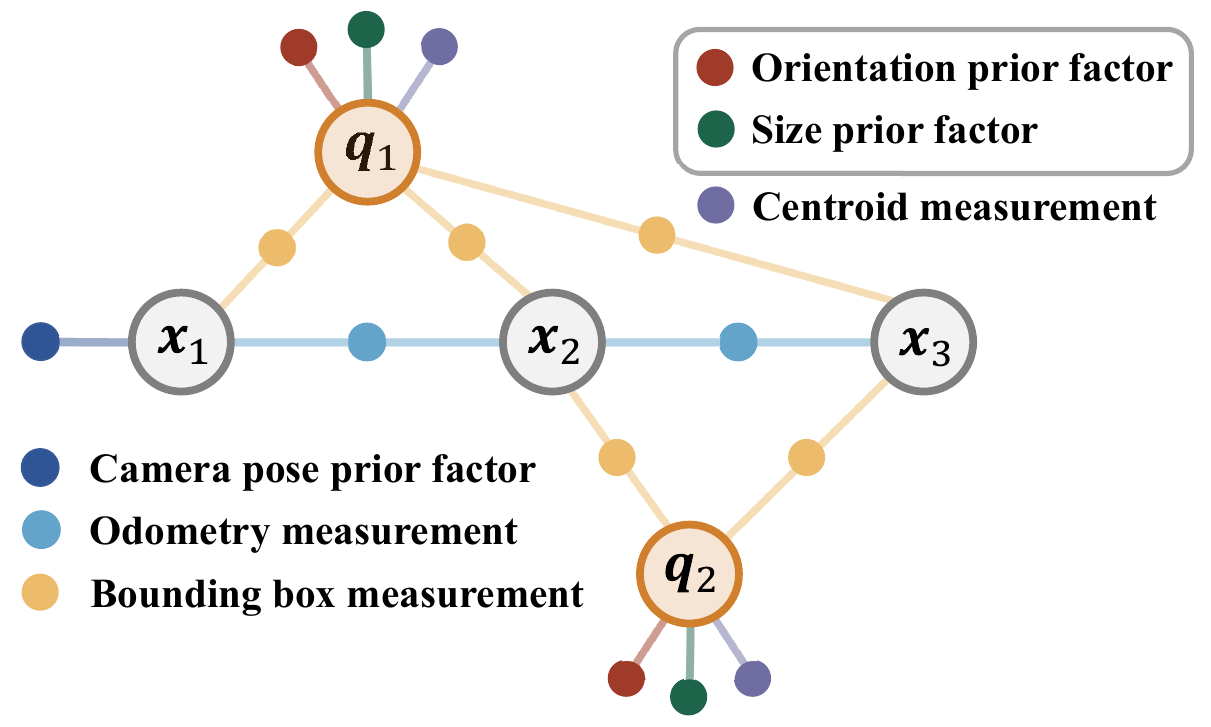}}
    \caption{A toy example to show the factor graph structure in our object SLAM. Object orientation and size prior factors are encoded with commonsense knowledge from LLM.}
    \label{fig:factor-graph}
\end{figure}

%% file: table/tum-map.tex
% TUM RGBD
\begin{center}
\setlength\tabcolsep{5pt}
% \begin{table}[t]

% \centering
% \caption{The object mapping results on the TUM-RGBD dataset}
% \label{table:tum-map}
%     \begin{tabular}{l cc|cc|cc}
    
% \toprule[1pt]
%   & 
%   \multicolumn{2}{c|}{\makecell{3D IoU $\uparrow $}} & 
%   \multicolumn{2}{c|}{\makecell{$\epsilon_d$ (m) $\downarrow $}} & 
%   \multicolumn{2}{c}{\makecell{$\epsilon_s$ (m) $\downarrow $}} \\ 
%          &  EAO  & Ours  & EAO & Ours & EAO & Ours \\
%      \midrule
%      fr1\textunderscore desk1  & 0.125 & \textbf{0.184} & 0.097 & \textbf{0.078} &  0.092 & \textbf{0.078} \\
%      fr2\textunderscore desk   & 0.161 & \textbf{0.396} & 0.141 & \textbf{0.050} & 0.073 & \textbf{0.038} \\
%      fr3\textunderscore office & 0.115 &\textbf{0.299}  & 0.136 & \textbf{0.052} & 0.057 & \textbf{0.044} \\
% \bottomrule[1pt]
%     \end{tabular}
% \end{table}

% ATE ORB3     EAO      Ours
% fr1 0.017676 0.014198 0.018217
% fr2 0.010976 0.010398 0.032721
% fr3 0.010686 0.031475 0.020230

\begin{table*}[t]
\centering
\caption{The object mapping and localization results on the TUM-RGBD dataset}
\label{table:tum-map}
\setlength\tabcolsep{3pt}
\begin{tabular}{l cccc|cccc|cccc|cccc|c}
\toprule[1pt]
\multirow{2}{*}{Seq.} & 
\multicolumn{4}{c|}{3D IoU $\uparrow$} & 
\multicolumn{4}{c|}{Centroid (m) $\downarrow$} & 
\multicolumn{4}{c|}{Size (m) $\downarrow$} &
\multicolumn{5}{c}{ATE (cm) $\downarrow$} \\ 
\cmidrule(lr){2-5} \cmidrule(lr){6-9} \cmidrule(lr){10-13} \cmidrule(lr){14-18}
& EAO & VOOM & Ours & Ours+W & EAO & VOOM & Ours & Ours+W & EAO & VOOM & Ours & Ours+W & EAO & VOOM & Ours & Ours+W & ORB3 \\
 % & E & V & O & O+W & E & V & O & O+W & E & V & O & O+W & E & V & O & O+W & ORB3 \\
\midrule
fr1 & 0.125 & - & \underline{0.184} & \textbf{0.242} & 0.097 & - & \underline{0.078} & \textbf{0.077} & 0.092 & - & \underline{0.078} & \textbf{0.054} & \textbf{1.4894} & - & 1.8217 & \underline{1.7678} & 1.7676\\
fr2 & 0.212 & 0.226 & \underline{0.338} & \textbf{0.378} & 0.170 & 0.096 & \underline{0.080} & \textbf{0.062} & 0.147 & \underline{0.049} & 0.055 & \textbf{0.047} & 1.2878 & 2.3245 & \underline{1.1084} & \textbf{1.0952} & 1.0976 \\
fr3 & 0.127 & 0.244 & \underline{0.305} & \textbf{0.326} & 0.216 & 0.064 & \underline{0.062} & \textbf{0.048} & 0.094 & \underline{0.044} & 0.046 & \textbf{0.041} & 1.3077 & 1.8233 & \underline{1.0740} & \textbf{1.0663} & 1.0686 \\
\bottomrule[1pt]
\end{tabular}
\end{table*}

% \begin{table}[t]
% \centering
% \caption{Object mapping results on TUM-RGBD dataset}
% \label{table:tum-map}
% \small
% \setlength\tabcolsep{1.5pt}
% \begin{tabular}{l c c c c | c c c c | c c c c}
% \toprule[0.8pt]
% \multirow{2}{*}{\rotatebox{90}{Seq}} & 
% \multicolumn{4}{c|}{IoU $\uparrow$} & 
% \multicolumn{4}{c|}{Cent.~(m) $\downarrow$} & 
% \multicolumn{4}{c}{Size~(m) $\downarrow$} \\ 
% \cmidrule(lr){2-5} \cmidrule(lr){6-9} \cmidrule(lr){10-13}
%  & E & V & O & O+W & E & V & O & O+W & E & V & O & O+W \\
% \midrule
% fr1 & .125 & - & \textbf{.184} & - & .097 & - & \textbf{.078} & - & .092 & - & \textbf{.078} & - \\
% fr2 & .212 & .226 & .338 & \textbf{.378} & .170 & .096 & .080 & \textbf{.062} & .147 & .049 & .055 & \textbf{.047} \\
% fr3 & .127 & .244 & .305 & \textbf{.326} & .216 & .064 & .062 & \textbf{.048} & .094 & .044 & .046 & \textbf{.041} \\
% \bottomrule[0.8pt]
% \end{tabular}
% \end{table}
\end{center}

%% file: table/tum-association.tex
% TUM RGBD
\begin{center}
\setlength{\belowcaptionskip}{-10pt} 
\setlength{\textfloatsep}{0pt} 
\setlength\tabcolsep{4.5pt}
\begin{table}[t]

\centering
\caption{The object association results on the TUM-RGBD dataset}
\label{table:tum-association}
    \begin{tabular}{l cccc|cccc|cccc|c}
    
\toprule[1pt]
   Seq. & 
  \multicolumn{4}{c|}{\makecell{TP}} & 
  \multicolumn{4}{c|}{\makecell{FP}} & 
  \multicolumn{4}{c|}{\makecell{FN}} &
  Total
  % \multicolumn{3}{c}{\makecell{Total Object Count}} 
  \\ 
     \midrule
         &  E & V & O & W & E & V & O & W & E & V & O & W & GT\\
     fr1  & 8 & - & \textbf{13} & 12 & 1 & - & \textbf{0} & \textbf{0} & 7 & - & \textbf{2} & 3 & 15 \\
     fr2   & 27 & \textbf{29} & 23 & 25 & \textbf{3} & 17 & 6 & 4 & 2 & \textbf{0} & 6 & 4 & 29 \\
     fr3 & 36 & \textbf{44} & 40 & 42 & \textbf{0} & 38 & 16 & 7 & 11 & \textbf{3} & 7 & 5 & 47 \\
\bottomrule[1pt]
    \end{tabular}
\end{table}
% \vspace{-10pt}
\end{center}

%% file: figure_latex/tum_incremental.tex
\begin{figure}[tb]
    \centering
    \begin{subfigure}{0.25\textwidth}
        \centering
        \includegraphics[width=0.99\textwidth]{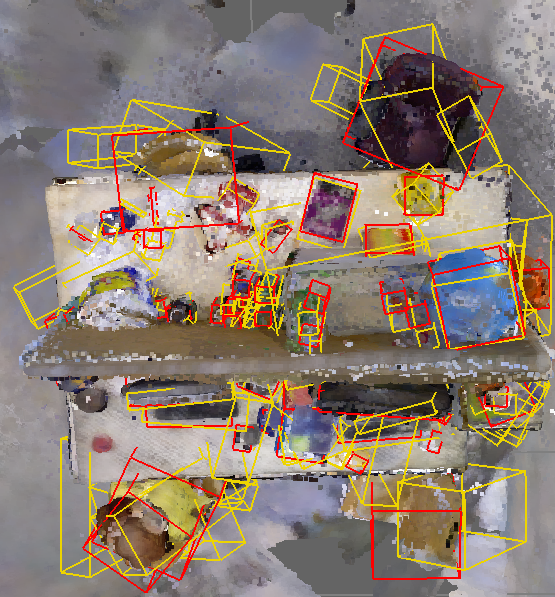}
        \caption{VOOM}
    \end{subfigure}%
    \begin{subfigure}{0.25\textwidth}
        \centering
        \includegraphics[width=0.94\textwidth]{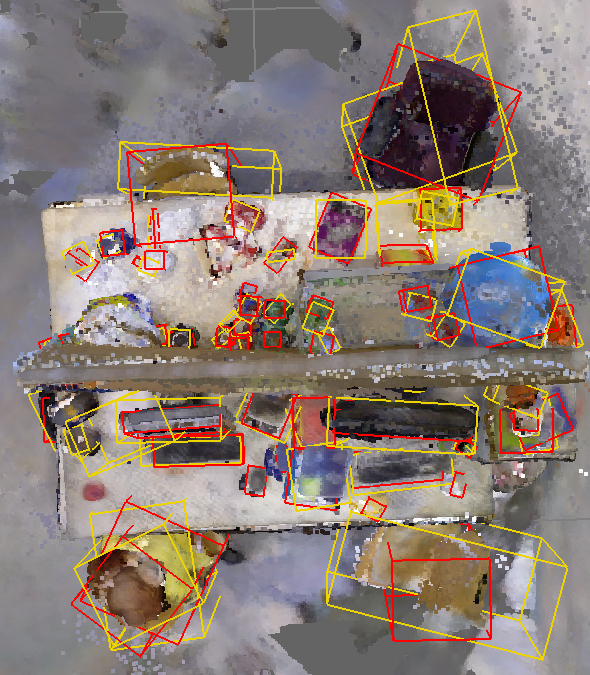}
        \caption{Our SLAM}
    \end{subfigure}%

    % \begin{subfigure}{0.19\textwidth}
    %     \centering
    %     \includegraphics[width=0.99\textwidth]{figure/ir1.png}
    %     \caption{t = 5s}
    % \end{subfigure}%
    % \begin{subfigure}{0.19\textwidth}
    %     \centering
    %     \includegraphics[width=0.99\textwidth]{figure/ir2.png}
    %     \caption{t = 19.6s}
    % \end{subfigure}%
    % \begin{subfigure}{0.19\textwidth}
    %     \centering
    %     \includegraphics[width=0.99\textwidth]{figure/ir3.png}
    %     \caption{t = 35.9s}
    % \end{subfigure}%
    % \begin{subfigure}{0.19\textwidth}
    %     \centering
    %     \includegraphics[width=0.99\textwidth]{figure/ir4.png}
    %     \caption{t = 86.8s}
    %     \label{subfig:tum-prior-final}
    % \end{subfigure}
    % \begin{subfigure}{0.22\textwidth}
    %     \centering
    %     \includegraphics[width=0.85\textwidth]{figure/fr2_no_p.png}
    %     \caption{object map w/o prior}
    %     \label{subfig:tum-no-prior}
    % \end{subfigure}
    \caption{
    The quantitative object mapping results on \textit{fr3\textunderscore office} using VOOM and our SLAM method. Quadrics are converted to cubes (yellow) to compare with labeled ground-truth objects (red). 
    % The online estimation of quadric objects on the TUM RGBD fr2\textunderscore desk dataset. We show the mapping procedure from (a) to (d) using our data association method with prior factors. In (e) we show the mapping result of the data association without prior factors. 
    }
    \label{fig:tum}
\end{figure}

%% file: figure_latex/tum_iou.tex
\begin{figure}[tb]
    \setlength{\belowcaptionskip}{-10pt} 
    % \centering
    \centerline{\includegraphics[width=0.48\textwidth]{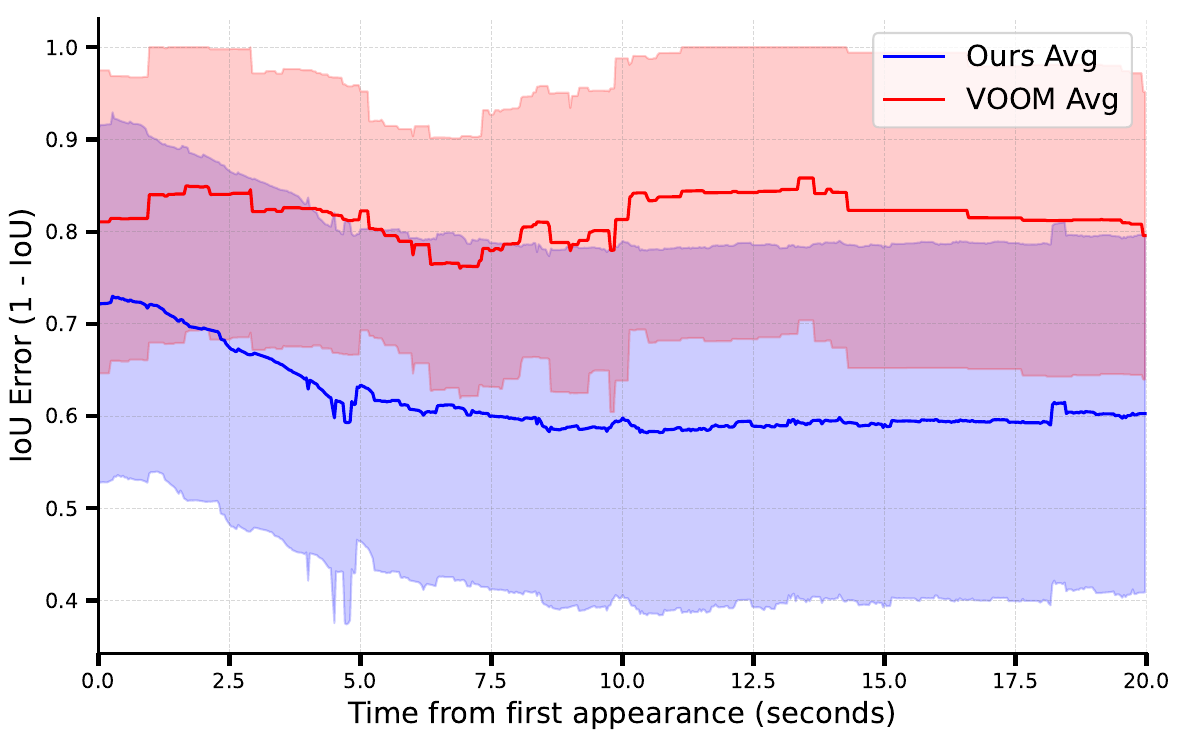}}
    \caption{Average 3D IoU error comparison between VOOM and our SLAM. Background colors of red and blue indicate the standard deviation of IoU error among objects for each method, where the purple area is simply the overlapping. The experiment was conducted on \textit{fr2\textunderscore desk} with YOLOv8x-seg detector.}
    \label{fig:tum-iou}
\end{figure}

%% file: figure_latex/mapping_3RScan.tex
\begin{figure*}[!tbp]
    \setlength{\belowcaptionskip}{-6pt} 
    \centering
    \begin{subfigure}{0.22\textwidth}
        \centering
        \includegraphics[width=0.9\textwidth]{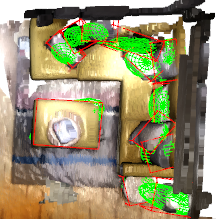}
        \caption{Living Room 1}
    \end{subfigure}%
    \begin{subfigure}{0.22\textwidth}
        \centering
        \includegraphics[width=0.9\textwidth]{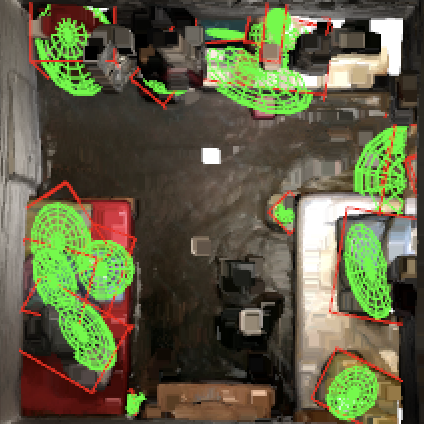}
        \caption{Bedroom 1}
    \end{subfigure}%
    \begin{subfigure}{0.22\textwidth}
        \centering
        \includegraphics[width=0.9\textwidth]{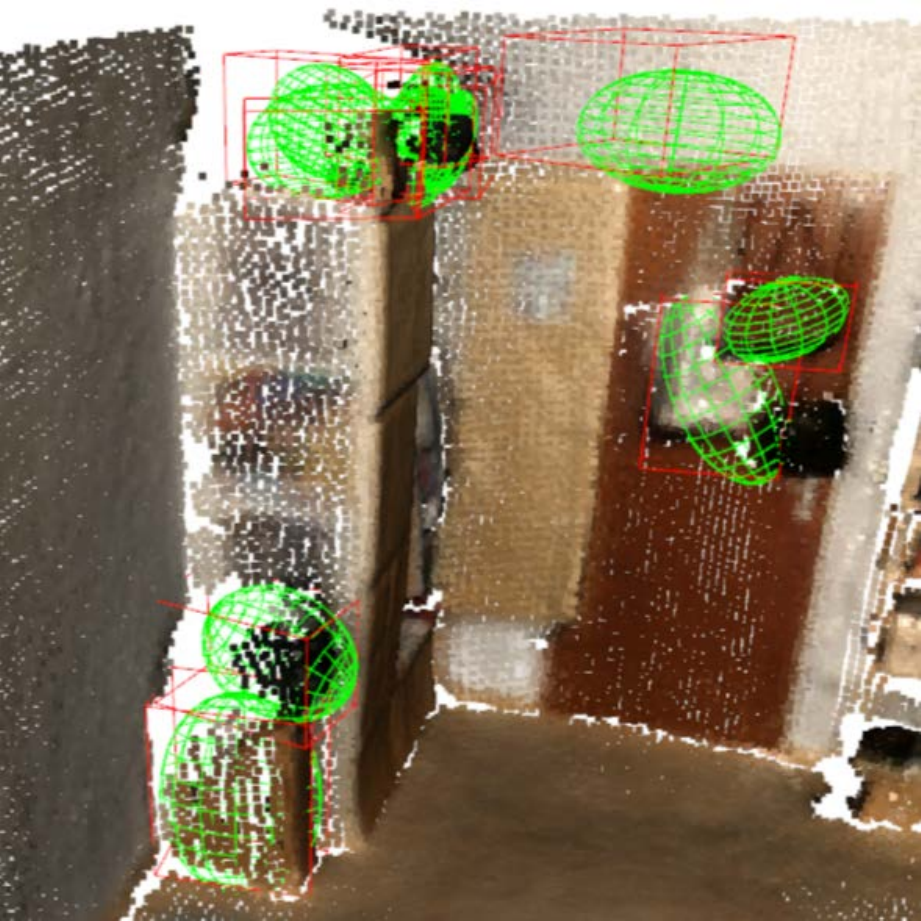}
        \caption{Bedroom 2}
        \label{subfig:3rscan-bedroom2}
    \end{subfigure}%
    \begin{subfigure}{0.22\textwidth}
        \centering
        \includegraphics[width=0.9\textwidth]{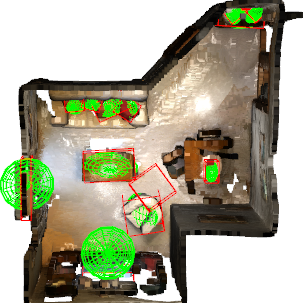}
        \caption{Living Room 2}
        \label{subfig:3rscan-living2}
    \end{subfigure}
    
    \caption{The final object mapping results on the 3RScan dataset. 
    The ellipsoids (green) represent the estimated object landmarks, and the cubes (red) are the object 3D bounding boxes from point cloud segmentation. Note that there are some bounding boxes without associated point cloud showing in the figure. These are objects located on the ceiling, and we cropped the point cloud for the ceiling to have a better visualization effect.}
    \label{fig:3RScan}
\end{figure*}

%% file: table/3RScan.tex
% (a) Offline w/ init & w/o prior  
% 0.0852860477161401  1.3015769393694885  5.771384227381051 (0.09667342349968046 
% if exclude one extremely large cost in the last dataset)
% (b) Offline w/o init & w/o prior 0.17101086490473155 2.570059436669362 2.579737945306944
% (c) Offline w/ init & w/ prior 0.081747732679329 1.3167916654027247 0.32280406664108086 (again 0.09004963655092622 if exclude last dataset)
% (d) Offline w/o init & w/ prior 0.10911069120508336 1.9355709340643226 1.9355709340643226

% & baseline & improved initialization & batch optimization & our method

\begin{center}
\setlength{\textfloatsep}{0pt} 
\setlength\tabcolsep{5pt}
\begin{table}
\centering
\caption{The experiment results on the 3RScan dataset}
\label{table:3rscan}
\begin{tabular}{ c c c  | c c }
\toprule[1pt]
  \multirow{2}{*}{} & 
  \multicolumn{2}{c|}{\makecell{Online Update}} & 
  \multicolumn{2}{c}{\makecell{Batch Optimization}}\\ 
% \hline
        & w/o priors  & Ours  & w/o priors & Ours \\
\midrule
3D IoU  $\uparrow $ & 0.320 & \textbf{0.430} & 0.400 & \textbf{0.423} \\  
Centroid (m) $\downarrow $ & 0.223  & \textbf{0.090} & 0.148 & \textbf{0.086} \\  
Size (m) $\downarrow $ & 0.318  & \textbf{0.095} & 0.196 & \textbf{0.094}  \\
 \bottomrule[1pt]
\end{tabular} 
\vspace{-2pt}

% $(\text{m})$ $(\text{m}/\text{m}^3)$
% \footnotesize{$^1$ Estimation with naive initialization and without contextual prior;\\ $^2$ Estimation with filtered initialization and contextual prior.}

\end{table}
\end{center}

% online-quadric
% error:  0.2228849360637064
% error norm:  3.9539896366289895
% size error:  0.3176355334058017
% cost:  2.048642626491048

% online-init
% error:  0.15174079317643274
% error norm:  2.7310723721631516
% size error:  0.11657568626665514
% cost:  0.4179188803354972

% online-ours
% error:  0.09003160586698525
% error norm:  1.558107755628533
% size error:  0.09522005083949853
% cost:  0.2891853181272669

% batch-quadricslam
% error:  0.14753559803497923
% error norm:  2.5319591563344703
% size error:  0.1960684831929041
% cost:  0.45827881373241297

% bacth-init
% error:  0.08579718104362802
% error norm:  1.3394797646112655
% size error:  0.1072906554138715
% cost:  0.024261101450969363

% batch-ours
% error:  0.08577736103516717
% error norm:  1.4868553693246802
% size error:  0.09365322865948514
% cost:  0.033473434223108334

%% file: table/ablation.tex
% TUM RGBD Ablation
\begin{center}
\setlength{\belowcaptionskip}{-6pt} 
\setlength{\textfloatsep}{0pt} 
\setlength\tabcolsep{5pt}

\begin{table}[t]
\centering
\caption{Ablation study of LLM-enhanced prior factors on the TUM-RGBD dataset}
\label{table:ablation}
\setlength\tabcolsep{2.5pt}

% \begin{tabular}{l ccc|ccc|ccc}
% \toprule[1pt]
% \multirow{2}{*}{Seq.} & 
% \multicolumn{3}{c|}{3D IoU $\uparrow$} & 
% \multicolumn{3}{c|}{Centroid (m) $\downarrow$} & 
% \multicolumn{3}{c}{Size (m) $\downarrow$} \\ 
% \cmidrule(lr){2-4} \cmidrule(lr){5-7} \cmidrule(lr){8-10} 
% & None & Ori & Size & None & Ori & Size & None & Ori & Size \\
%  % & E & V & O & O+W & E & V & O & O+W & E & V & O & O+W & E & V & O & O+W & ORB3 \\
% \midrule
% fr2 & 0.212 & 0.226 & \underline{0.338} & \textbf{0.378} & 0.170 & 0.096 & \underline{0.080} & \textbf{0.062} & 0.147 \\
% fr3 & 0.127 & 0.244 & \underline{0.305} & \textbf{0.326} & 0.216 & 0.064 & \underline{0.062} & \textbf{0.048} & 0.094\\
% \bottomrule[1pt]
% \end{tabular}

\begin{tabular}{l cccc|cccc|cccc}
\toprule[1pt]
\multirow{2}{*}{Seq.} & 
\multicolumn{4}{c|}{3D IoU $\uparrow$} & 
\multicolumn{4}{c|}{Centroid (m) $\downarrow$} & 
\multicolumn{4}{c}{Size (m) $\downarrow$} \\ 
\cmidrule(lr){2-5} \cmidrule(lr){6-9} \cmidrule(lr){10-13} 
% & None & Ori & Size & Full & None & Ori & Size & Full & None & Ori & Size & Full \\
& N & O & S & F & N & O & S & F & N & O & S & F \\
 % & E & V & O & O+W & E & V & O & O+W & E & V & O & O+W & E & V & O & O+W & ORB3 \\
% .235 & .257 & .234 & 0.326 & .069 & .064 & .068 & 0.048 & .080 & .081 & .080 & 0.041

\midrule
fr2 & .275 & .316 & .292 & \textbf{.378} & .109 & .129 & .086 & \textbf{.062} & .073 & .081 & .063 & \textbf{.047} \\
fr3 & .235 & .257 & .234 & \textbf{.326} & .069 & .064 & .068 & \textbf{.048} & .080 & .081 & .080 & \textbf{.041} \\
\bottomrule[1pt]
\end{tabular}

\end{table}
\end{center}

% \setlength\tabcolsep{5pt}
% \renewcommand{\arraystretch}{1.2}
% \begin{table}[t]
%     \caption{Ablation}
%     \centering
%     \begin{tabular}{cccc|cc}
%          & size& centroid & orientation & $\epsilon_f$& $\bar{\epsilon_f}$ \\[2pt]
%          1& \cmark &\cmark&\cmark&
%          \textbf{10798.78}& \textbf{0.783} \\ 
%          2& \cmark &  \xmark&    \cmark        &
%          11859.87& 0.862\\
%          3& \xmark &\cmark&\cmark&
%          33030.80& 2.400 \\
         
%          4& \cmark&\cmark&\xmark&
%          35741.75& 2.597 \\
%          % 5&            & \cmark   &   & 34009.35 & 2.477 & \\
%          % 6& \cmark &  &   & 35741.75 & 2.604 & \\
%          % 7&            &   & \cmark  & 32508.74 & 2.368 & \\
%          5&      \xmark      &  \xmark  & \xmark
%          & 34009.35 & 2.484 \\
%     \end{tabular}
%     \label{table:ablation}
% \end{table}

%% file: main.bbl
\begin{thebibliography}{10}
\providecommand{\url}[1]{#1}
\csname url@samestyle\endcsname
\providecommand{\newblock}{\relax}
\providecommand{\bibinfo}[2]{#2}
\providecommand{\BIBentrySTDinterwordspacing}{\spaceskip=0pt\relax}
\providecommand{\BIBentryALTinterwordstretchfactor}{4}
\providecommand{\BIBentryALTinterwordspacing}{\spaceskip=\fontdimen2\font plus
\BIBentryALTinterwordstretchfactor\fontdimen3\font minus \fontdimen4\font\relax}
\providecommand{\BIBforeignlanguage}[2]{{%
\expandafter\ifx\csname l@#1\endcsname\relax
\typeout{** WARNING: IEEEtran.bst: No hyphenation pattern has been}%
\typeout{** loaded for the language `#1'. Using the pattern for}%
\typeout{** the default language instead.}%
\else
\language=\csname l@#1\endcsname
\fi
#2}}
\providecommand{\BIBdecl}{\relax}
\BIBdecl

\bibitem{wu2023object}
Y.~Wu, Y.~Zhang, D.~Zhu, Z.~Deng, W.~Sun, X.~Chen, and J.~Zhang, ``An object slam framework for association, mapping, and high-level tasks,'' \emph{IEEE Transactions on Robotics}, vol.~39, no.~4, pp. 2912--2932, 2023.

\bibitem{zhou2023efficient}
H.~Zhou, Z.~Hu, S.~Liu, and S.~Khan, ``Efficient 2d graph slam for sparse sensing,'' \emph{arXiv preprint arXiv:2312.02353}, 2023.

\bibitem{jablonsky2018orientation}
N.~Jablonsky, M.~Milford, and N.~S{\"u}nderhauf, ``An orientation factor for object-oriented slam,'' \emph{arXiv preprint arXiv:1809.06977}, 2018.

\bibitem{ge2024commonsense}
W.~Ge, C.~Tang, and H.~Zhang, ``Commonsense scene graph-based target localization for object search,'' in \emph{2024 IEEE/RSJ International Conference on Intelligent Robots and Systems (IROS)}.\hskip 1em plus 0.5em minus 0.4em\relax IEEE, 2024, pp. 13\,318--13\,325.

\bibitem{factorgraph910572}
F.~Kschischang, B.~Frey, and H.-A. Loeliger, ``Factor graphs and the sum-product algorithm,'' \emph{IEEE Transactions on Information Theory}, vol.~47, no.~2, pp. 498--519, 2001.

\bibitem{hosseinzadeh2019real}
M.~Hosseinzadeh, K.~Li, Y.~Latif, and I.~Reid, ``Real-time monocular object-model aware sparse slam,'' in \emph{2019 international conference on robotics and automation (ICRA)}.\hskip 1em plus 0.5em minus 0.4em\relax IEEE, 2019, pp. 7123--7129.

\bibitem{salas2013slam++}
R.~F. Salas-Moreno, R.~A. Newcombe, H.~Strasdat, P.~H. Kelly, and A.~J. Davison, ``Slam++: Simultaneous localisation and mapping at the level of objects,'' in \emph{Proceedings of the IEEE conference on computer vision and pattern recognition}, 2013, pp. 1352--1359.

\bibitem{nicholson2018quadricslam}
L.~Nicholson, M.~Milford, and N.~S{\"u}nderhauf, ``Quadricslam: Dual quadrics from object detections as landmarks in object-oriented {SLAM},'' \emph{IEEE Robotics and Automation Letters}, vol.~4, no.~1, pp. 1--8, 2018.

\bibitem{yang2019cubeslam}
S.~Yang and S.~Scherer, ``Cubeslam: Monocular 3-d object slam,'' \emph{IEEE Transactions on Robotics}, vol.~35, no.~4, pp. 925--938, 2019.

\bibitem{wu2020eao}
Y.~Wu, Y.~Zhang, D.~Zhu, Y.~Feng, S.~Coleman, and D.~Kerr, ``Eao-slam: Monocular semi-dense object slam based on ensemble data association,'' in \emph{2020 IEEE/RSJ International Conference on Intelligent Robots and Systems (IROS)}.\hskip 1em plus 0.5em minus 0.4em\relax IEEE, 2020, pp. 4966--4973.

\bibitem{liao2022so}
Z.~Liao, Y.~Hu, J.~Zhang, X.~Qi, X.~Zhang, and W.~Wang, ``{So-slam}: Semantic object {SLAM} with scale proportional and symmetrical texture constraints,'' \emph{IEEE Robotics and Automation Letters}, vol.~7, no.~2, pp. 4008--4015, 2022.

\bibitem{wang2024voom}
Y.~Wang, C.~Jiang, and X.~Chen, ``Voom: Robust visual object odometry and mapping using hierarchical landmarks,'' in \emph{2024 IEEE International Conference on Robotics and Automation (ICRA)}.\hskip 1em plus 0.5em minus 0.4em\relax IEEE, 2024, pp. 10\,298--10\,304.

\bibitem{qian2021semantic}
Z.~Qian, K.~Patath, J.~Fu, and J.~Xiao, ``Semantic slam with autonomous object-level data association,'' in \emph{2021 IEEE International Conference on Robotics and Automation (ICRA)}.\hskip 1em plus 0.5em minus 0.4em\relax IEEE, 2021, pp. 11\,203--11\,209.

\bibitem{zins2022oa}
M.~Zins, G.~Simon, and M.-O. Berger, ``Oa-slam: Leveraging objects for camera relocalization in visual slam,'' in \emph{2022 IEEE international symposium on mixed and augmented reality (ISMAR)}.\hskip 1em plus 0.5em minus 0.4em\relax IEEE, 2022, pp. 720--728.

\bibitem{fang2017object}
Y.~Fang, K.~Kuan, J.~Lin, C.~Tan, and V.~Chandrasekhar, ``Object detection meets knowledge graphs,'' in \emph{Proceedings of the 26th International Joint Conference on Artificial Intelligence}, ser. IJCAI'17.\hskip 1em plus 0.5em minus 0.4em\relax AAAI Press, 2017, p. 1661–1667.

\bibitem{pal2021learning}
A.~Pal, Y.~Qiu, and H.~Christensen, ``Learning hierarchical relationships for object-goal navigation,'' in \emph{Conference on Robot Learning}.\hskip 1em plus 0.5em minus 0.4em\relax PMLR, 2021, pp. 517--528.

\bibitem{hosseinzadeh2019structure}
M.~Hosseinzadeh, Y.~Latif, T.~Pham, N.~Suenderhauf, and I.~Reid, ``Structure aware {SLAM} using quadrics and planes,'' in \emph{Computer Vision--ACCV 2018: 14th Asian Conference on Computer Vision, Perth, Australia, December 2--6, 2018, Revised Selected Papers, Part III 14}.\hskip 1em plus 0.5em minus 0.4em\relax Springer, 2019, pp. 410--426.

\bibitem{shah2023navigation}
D.~Shah, M.~R. Equi, B.~Osi{\'n}ski, F.~Xia, B.~Ichter, and S.~Levine, ``Navigation with large language models: Semantic guesswork as a heuristic for planning,'' in \emph{Conference on Robot Learning}.\hskip 1em plus 0.5em minus 0.4em\relax PMLR, 2023, pp. 2683--2699.

\bibitem{zhao2023large}
Z.~Zhao, W.~S. Lee, and D.~Hsu, ``Large language models as commonsense knowledge for large-scale task planning,'' \emph{Advances in Neural Information Processing Systems}, vol.~36, pp. 31\,967--31\,987, 2023.

\bibitem{yang2024thinking}
J.~Yang, S.~Yang, A.~W. Gupta, R.~Han, L.~Fei-Fei, and S.~Xie, ``Thinking in space: How multimodal large language models see, remember, and recall spaces,'' \emph{arXiv preprint arXiv:2412.14171}, 2024.

\bibitem{kaess2012isam2}
M.~Kaess, H.~Johannsson, R.~Roberts, V.~Ila, J.~J. Leonard, and F.~Dellaert, ``{iSAM2}: Incremental smoothing and mapping using the bayes tree,'' \emph{The International Journal of Robotics Research}, vol.~31, no.~2, pp. 216--235, 2012.

\bibitem{Jocher_Ultralytics_YOLO_2023}
\BIBentryALTinterwordspacing
G.~Jocher, A.~Chaurasia, and J.~Qiu, ``Ultralytics yolo,'' Jan. 2023, version 8.0.0, AGPL-3.0 license. [Online]. Available: \url{https://github.com/ultralytics/ultralytics}
\BIBentrySTDinterwordspacing

\bibitem{sort8296962}
N.~Wojke, A.~Bewley, and D.~Paulus, ``Simple online and realtime tracking with a deep association metric,'' in \emph{2017 IEEE International Conference on Image Processing (ICIP)}, 2017, pp. 3645--3649.

\bibitem{deepsort8296962}
------, ``Simple online and realtime tracking with a deep association metric,'' in \emph{2017 IEEE International Conference on Image Processing (ICIP)}, 2017, pp. 3645--3649.

\bibitem{aharon2022botsort}
N.~Aharon, R.~Orfaig, and B.-Z. Bobrovsky, ``{BoT-SORT}: Robust associations multi-pedestrian tracking,'' 2022.

\bibitem{pennington2014glove}
\BIBentryALTinterwordspacing
J.~Pennington, R.~Socher, and C.~D. Manning, ``Glove: Global vectors for word representation,'' in \emph{Empirical Methods in Natural Language Processing (EMNLP)}, 2014, pp. 1532--1543. [Online]. Available: \url{http://www.aclweb.org/anthology/D14-1162}
\BIBentrySTDinterwordspacing

\bibitem{tum2012rgbd-dataset}
J.~Sturm, N.~Engelhard, F.~Endres, W.~Burgard, and D.~Cremers, ``A benchmark for the evaluation of {RGB-D SLAM} systems,'' in \emph{2012 IEEE/RSJ International Conference on Intelligent Robots and Systems}, 2012, pp. 573--580.

\bibitem{Wald2019RIO}
J.~Wald, A.~Avetisyan, N.~Navab, F.~Tombari, and M.~Niessner, ``{RIO: 3D Object Instance Re-Localization in Changing Indoor Environments},'' in \emph{Proceedings IEEE International Conference on Computer Vision (ICCV)}, 2019.

\bibitem{Campos_2021orb3}
\BIBentryALTinterwordspacing
C.~Campos, R.~Elvira, J.~J.~G. Rodriguez, J.~M. M.~Montiel, and J.~D.~Tardos, ``{ORB-SLAM3}: An accurate open-source library for visual, visual–inertial, and multimap {SLAM},'' \emph{IEEE Transactions on Robotics}, vol.~37, no.~6, p. 1874–1890, Dec. 2021. [Online]. Available: \url{http://dx.doi.org/10.1109/TRO.2021.3075644}
\BIBentrySTDinterwordspacing

\bibitem{sager2021labelcloud}
C.~Sager, P.~Zschech, and N.~Kühl, ``labelcloud: A lightweight domain-independent labeling tool for 3d object detection in point clouds,'' 2021.

\bibitem{Cheng2024YOLOWorld}
T.~Cheng, L.~Song, Y.~Ge, W.~Liu, X.~Wang, and Y.~Shan, ``Yolo-world: Real-time open-vocabulary object detection,'' in \emph{Proc. IEEE Conf. Computer Vision and Pattern Recognition (CVPR)}, 2024.

\bibitem{ravi2020pytorch3d}
N.~Ravi, J.~Reizenstein, D.~Novotny, T.~Gordon, W.-Y. Lo, J.~Johnson, and G.~Gkioxari, ``Accelerating 3d deep learning with pytorch3d,'' \emph{arXiv:2007.08501}, 2020.

\end{thebibliography}
